\documentclass[a4paper,fleqn]{cas-dc}

\usepackage[numbers]{natbib}

\usepackage{graphicx}
\usepackage{subfigure}
\usepackage{array}

\usepackage{float}
\usepackage{booktabs}
\usepackage{bbding}
\usepackage{etoolbox}
\usepackage{color}
\usepackage{soul}

\usepackage{algorithm}
\usepackage{algorithmicx}

\usepackage{algpseudocode}
\usepackage{balance}
\usepackage{amsmath, amssymb}
\usepackage{makecell}
\usepackage{multirow}
\usepackage{hyperref}
\usepackage{tabu}
\def\tsc#1{\csdef{#1}{\textsc{\lowercase{#1}}\xspace}}
\tsc{WGM}
\tsc{QE}
\tsc{EP}
\tsc{PMS}
\tsc{BEC}
\tsc{DE}
\usepackage{soul}
\soulregister\ref{7}
\soulregister\cite{7}

\begin{document}
\let\WriteBookmarks\relax
\def\floatpagepagefraction{1}
\def\textpagefraction{.001}

\shorttitle{Rethinking Exemplars for Continual Semantic Segmentation in Endoscopy Scenes: Entropy-based Mini-Batch Pseudo-Replay}

\shortauthors{Guankun Wang et~al.}

\title [mode = title]{Rethinking Exemplars for Continual Semantic Segmentation in Endoscopy Scenes: Entropy-based Mini-Batch Pseudo-Replay}                      

\tnotemark[1]

\tnotetext[1]{This document is the results of the research projects by Hong Kong Research Grants Council (RGC) Research Impact Fund (RIF) R4020-22, Collaborative Research Fund (CRF C4026-21GF, CRF C4063-18G), General Research Fund (GRF 14203323),  NSFC/RGC Joint Research Scheme N\_CUHK420/22, GRS \#3110167; Shenzhen-Hong Kong-Macau Technology Research Programme (Type C) STIC Grant SGDX20210823103535014 (202108233000303); Guangdong Basic and Applied Basic Research Foundation (GBABF) \#2021B1515120035; Shun Hing Institute of Advanced Engineering (SHIAE Project BME-p1-21) at The Chinese University of Hong Kong (CUHK).}

\author[1]{Guankun Wang}[orcid=0000-0003-2440-4950]
\fnmark[1]

\ead{gkwang@link.cuhk.edu.hk}

\credit{Conceptualization, Methodology, Software, Validation, Formal analysis, Investigation, Data curation, Writing — Original draft preparation, Writing — Review and editing, Visualization}

\affiliation[1]{organization={Department of Electronic Engineering, The Chinese University of Hong Kong},
    city={Hong Kong},
    country={China}}

\author[1]{Long Bai}[orcid=0000-0002-9762-6821]
\fnmark[1]

\ead{b.long@ieee.org}
\credit{Conceptualization, Methodology, Software, Formal analysis, Investigation, Resources, Writing — Original draft preparation, Writing — Review and editing, Supervision, Project Administration}

\author[1,2,3]{Yanan Wu}[orcid=0000-0003-2291-1334]
\ead{yananwu@cuhk.edu.hk}

\credit{Methodology, Investigation, Writing - Review and editing}

\affiliation[2]{organization={College of Medicine and Biological Information Engineering, Northeastern University},
    city={Shenyang},
    country={China}}

\affiliation[3]{organization={Key Laboratory of Intelligent Computing in Medical Image, Ministry of Education, Northeastern University},
    city={Shenyang},
    country={China}}

\author[4]{Tong Chen}[orcid=0000-0003-4312-7151]

\ead{tche2095@uni.sydney.edu.au}

\credit{Writing - Original draft preparation}

\affiliation[4]{organization={School of Electrical and Information Engineering, The University of Sydney},
    city={Sydney},
    country={Australia}}

\author
[1,5,6,7,8]{Hongliang Ren}[orcid=0000-0002-6488-1551]
\cormark[1]

\ead{hlren@ee.cuhk.edu.hk}
\credit{Conceptualization, Resources, Writing — Review and editing, Supervision, Project Administration, Funding acquisition}

\affiliation[5]{organization={Department of Biomedical Engineering, National University of Singapore},
    country={Singapore}}
\affiliation[6]{organization={Suzhou Research Institute, National University of Singapore},
    city={Suzhou},
    country={China}}
\affiliation[7]{organization={Shun Hing Institute of Advanced Engineering, The Chinese University of Hong Kong},
    city={Hong Kong},
    country={China}}
\affiliation[8]{organization={Shenzhen Research Institute, The Chinese University of Hong Kong},
    city={Shenzhen},
    country={China}}

\cortext[cor1]{Corresponding author}

\fntext[fn1]{Guankun Wang and Long Bai are co-first authors as they have contributed equally to the manuscript.}

\begin{abstract}
Endoscopy is a widely used technique for the early detection of diseases or robotic-assisted minimally invasive surgery (RMIS). Numerous deep learning (DL)-based research works have been developed for automated diagnosis or processing of endoscopic view. However, existing DL models may suffer from catastrophic forgetting. When new target classes are introduced over time or cross institutions, the performance of old classes may suffer severe degradation. More seriously, data privacy and storage issues may lead to the unavailability of old data when updating the model. 
Therefore, it is necessary to develop a continual learning (CL) methodology to solve the problem of catastrophic forgetting in endoscopic image segmentation. To tackle this, we propose a \textbf{Endo}scopy \textbf{C}ontinual \textbf{S}emantic \textbf{S}egmentation (EndoCSS) framework that does not involve the storage and privacy issues of exemplar data. The framework includes a mini-batch pseudo-replay (MB-PR) mechanism and a self-adaptive noisy cross-entropy (SAN-CE) loss. The MB-PR strategy circumvents privacy and storage issues by generating pseudo-replay images through a generative model. Meanwhile, the MB-PR strategy can also correct the model deviation to the replay data and current training data, which is aroused by the significant difference in the amount of current and replay images. Therefore, the model can perform effective representation learning on both new and old tasks. SAN-CE loss can help model fitting by adjusting the model's output logits, and also improve the robustness of training. Extensive continual semantic segmentation (CSS) experiments on public datasets demonstrate that our method can robustly and effectively address the catastrophic forgetting brought by class increment in endoscopy scenes. The results show that our framework holds excellent potential for real-world deployment in a streaming learning manner.
\end{abstract}


\begin{keywords}
Continual learning  \sep Endoscopic view \sep Image segmentation \sep Image synthesis \sep Pseudo-replay
\end{keywords}

\maketitle

\section{Introduction}

\begin{figure*}
    \centering
    \includegraphics[width=0.95\textwidth, trim=0 0 0 0]{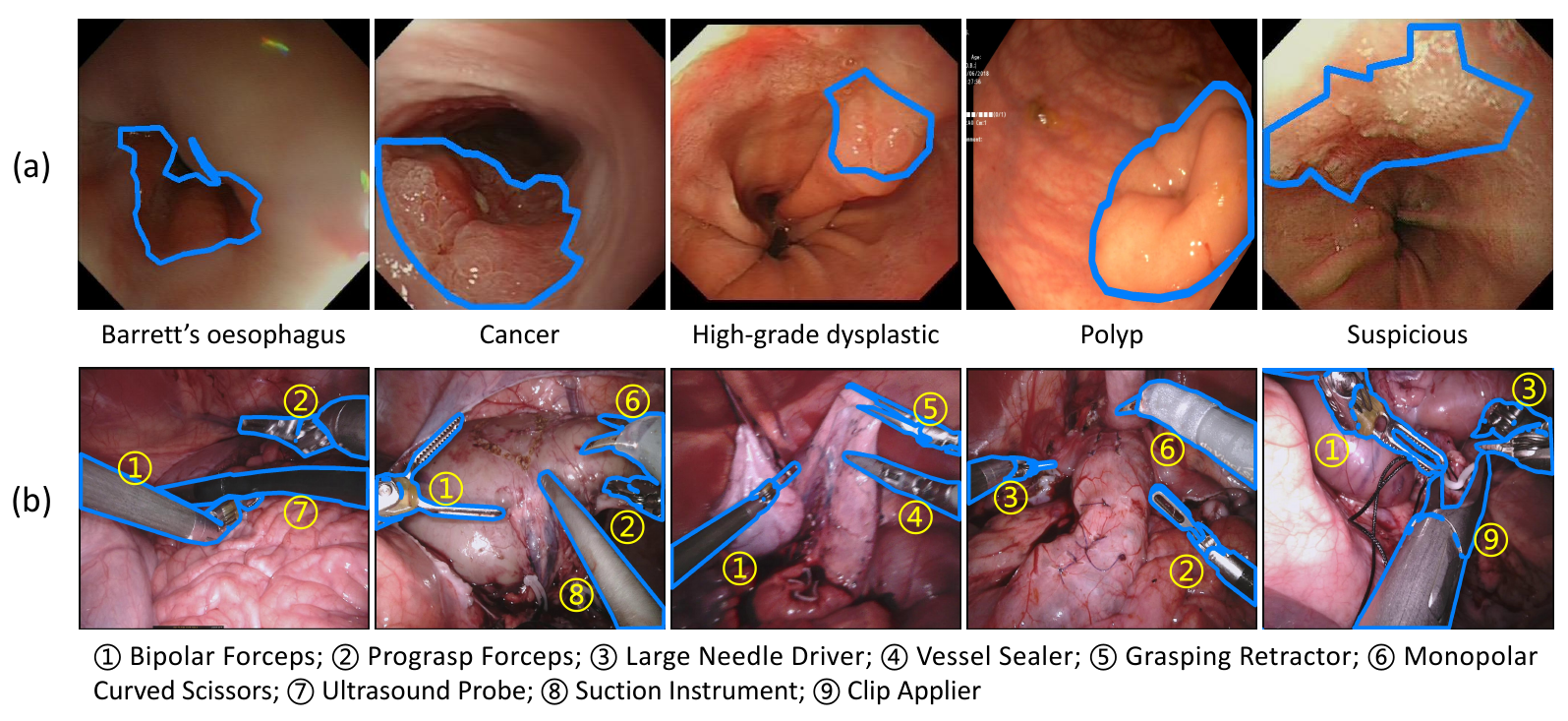}
    \caption{Endoscopic view examples: (a) EDD2020~\cite{ali2020endoscopy} dataset; (b) EndoVis18~\cite{allan2020endovis18} and EndoVis17~\cite{allan2019endovis17} datasets.}
    \label{fig:example}
\end{figure*}
 
Currently, endoscopy has been used as the gold standard for therapeutic procedures, early detection of diseases, and robotic-assisted minimally invasive surgery (RMIS)~\cite{le2020u}. During the endoscopy procedure, an endoscope is employed to visualize the interior of the target organs, and the signals will be presented on the external screen~\cite{ali2020endoscopy}, which helps the diagnosis, treatment, and curb the incidence of severe illness. However, the manual screening and annotation process can only be performed by experienced and well-trained clinicians. Furthermore, with the tremendous number of frames output by the endoscopy, the manual procedure will greatly prolong the time for the examination and annotation, causing a significant burden to the clinicians and more extended periods of suffering to the patients~\cite{xu2021reciprocally}. With the significant progress of deep learning (DL) and medical image analysis~\cite{bai2021influence,che2023towards,che2022learning,wu2023two}, DL-based endoscopy video diagnosis and treatment have been largely discussed and explored~\cite{jain2021deep,liu2023landmark,wang2023domain,zhang2022deep}. DL models can be trained with pre-labeled endoscopy data to improve efficiency and accuracy. As an effective auxiliary tool for disease diagnosis and treatment, DL models can provide clinicians with objective references, helping clinicians significantly save time and reduce burden~\cite{sushma2022recent}. Currently, DL-based endoscopy screening techniques have successfully tackled the application problems on different diseases, e.g., polyp~\cite{li2023semi, wang2023ra}, bleeding~\cite{jia2016deep, yuan2015automatic}, early gastric cancer~\cite{liu2018transfer}, Crohn’s disease~\cite{klang2020deep}, ulcer~\cite{masmoudi2022optimal}, etc.
In addition, the DL-based surgical scene segmentation and understanding tasks further promote the process of RMIS~\cite{islam2020learning, zia2023surgical}. With the help of DL models, the robot can understand the location and interaction of surgical tools and organs, and realize human-like analysis. As a result, the DL model can provide physicians with effective surgical skill assessment, monitoring, alerting, and post-operative analysis~\cite{seenivasan2022global}.

Nevertheless, deep neural networks (DNNs) cannot learn knowledge in a sequential setup~\cite{ammour2021lwfecg,che2023towards,sirshar2021incremental}. The performance of DNNs shall degrade dramatically on previous tasks when learning new ones, and this phenomenon is called \textit{catastrophic forgetting}~\cite{kirkpatrick2017overcoming}. Currently, few studies have explored the problem of catastrophic forgetting in endoscopic signals, especially the segmentation tasks requiring pixel-level accuracy. However, in the medical field, catastrophic forgetting often happens when we need to extend the DL models to new datasets with different classes or from other medical domains~\cite{baweja2018towards, gupta2021addressing}. More seriously, due to the strict restrictions on medical data privacy and licensing, the previous data may not be accessible when we refine the DL models on the new tasks~\cite{lee2020clinical}. Therefore, this paper addresses the class-incremental problem in endoscopy continual semantic segmentation (CSS). 
We focus on keeping updating our segmentation model with newly arrived data, and finalizing a self-adaptive DL model deployed in sequential and continual clinical applications. 
It is challenging for the model to distinguish similar features from different diseases~\cite{bai2022transformer} or instruments~\cite{bai2023revisiting} correctly. Figure~\ref{fig:example} visualize the sample data for gastrointestinal diseases and robotic-assisted surgery.

Therefore, we revisit the rehearsal strategies in continual learning (CL), and explore how to make the model conduct effective feature learning on the synthetic data. Firstly, synthetic medical data can be shared across the healthcare industry to facilitate knowledge discovery without revealing true patient-level data~\cite{nikolenko2021synthetic}, and the DL models shall be able to learn effectively from image samples generated from well-developed generative models.
We shall discuss three essential steps of the synthetic replay method: data selection, pseudo-label generation, and data feeding combination. The existing rehearsal method simply combines the replay and current data, but the current data usually accounts for a large proportion of the entire data set, and the model will have a significant bias on the current data. In this case, we propose our \textbf{Endo}scopy \textbf{C}ontinual \textbf{S}emantic \textbf{S}egmentation (EndoCSS) framework with the mini-batch pseudo-replay (MB-PR) strategy. The MB-PR strategy shall combine data at the batch level, and carefully sets the mini-batch level sample selection to increase the learning efficiency of the model. Furthermore, the framework employs our proposed self-adaptive noisy cross entropy (SAN-CE) loss to boost learning stability and performance. Overall, the contributions of this work can be summarized as follows:

\begin{itemize}
    \item[--] We propose a continual segmentation framework under the class increment for endoscopic view, providing valuable pioneering exploration and experience for continual learning research in the endoscopy. 
    \item[--] We provide an in-depth and detailed look at the synthetic replay data. We employ a mini-batch combination for the current and replay data, balancing the bias of the model on the replay data versus the current data when learning representations.
    \item[--] We adopt self-adaptive noisy cross-entropy (SAN-CE) loss to enhance the network optimization process, and improve the stability of the predictor to compensate for its bias in CL scenarios.
    \item[--] Extensive experiments demonstrate that our proposed framework can effectively deal with CSS problems in medical scenarios. Moreover, our method successfully solves the problem of privacy preservation, and opens up the possibility of deploying DL models in real-world streaming scenarios.
\end{itemize}

\section{Related Work}
\subsection{Continual learning in medical segmentation}

The application of CL techniques in medical image segmentation has been explored in a number of research works, mainly by adapting regularization and pseudo-rehearsal techniques. 
Researchers have investigated the capabilities of elastic weight consolidation (EWC)~\cite{kirkpatrick2017overcoming} method on the segmentation of white matter lesions~\cite{baweja2018towards} and glioma~\cite{van2019towards}. Besides, Özgün \textit{et al.}~\cite{ozgun2020importance} adapted the memory aware synapses (MAS) regularization~\cite{aljundi2018memory} to brain segmentation. Instead of the soft penalty, they directly restricted the adaptation of the important parameters.
Karani et al.~\cite{karani2018lifelong} introduced a U-Net with domain-specific batch normalization layers and shared convolutional layers. Their method mitigated the performance degradation of MRI brain segmentation caused by various scanners or scanning protocols. However, their method necessitated collecting all domain data at once~\cite{gonzalez2020wrong}. Furthermore, some research concentrates on learning feature transformations between feature spaces or domain-independent features. Elskhawy \textit{et al.}~\cite{elskhawy2020continual} separated domain-dependent features from domain-independent features through an adversarial technique, effectively achieving the balance of rigidity and plasticity in class-incremental learning setup. 

However, current medical CSS methods still revolve around techniques such as distillation and regularization to constrain and adjust model parameters. These existing methods mostly consider the implicit features in the model, as the limitations of data privacy and availability in medical scenarios prevent the model from freely accessing old data and fully utilizing explicit features. In this case, we propose to mitigate the impact of data privacy and availability by using a generative model, allowing the model to directly learn explicit representations from the original images. Especially, endoscopic images are predominantly RGB images. In disease or instrument segmentation tasks, the original RGB images offer a rich source of explicit features that are worth exploring across different classes. Therefore, we take into consideration the incorporation of more explicit features in our framework and devised a meticulously designed pseudo-rehearsal strategy. By leveraging the collaboration of explicit and implicit representations, we aim to effectively tackle the CSS problem in the context of the endoscopic view.

\subsection{CSS with rehearsal}
There are four primary (non-mutually exclusive) categories of current CL techniques~\cite{lesort2020continual}: dynamic architectures, regularization, rehearsal methods, and pseudo-replays. Rehearsal methods restore or generate data samples from previous tasks to preserve old knowledge. In CSS, Cha~\textit{et al.}~\cite{cha2021ssul} preserved a tiny exemplar set, and conducted joint training on the current and exemplar dataset. Nevertheless, it shall leak raw data into new tasks and raise concerns about data storage and privacy. The same issue shall also exist in~\cite{fortin2022continual}. RECALL~\cite{maracani2021recall} employed the generative model pre-trained on ImageNet~\cite{deng2009imagenet} and data grasped from the web to construct their exemplar set, while these two strategies are not applicable to rigorous medical problems. 
To tackle the above problems, our rehearsal mechanism will only contain the generative model, and not include any data storage. Entropy-based methods shall help us filter out poor-quality pairs of generated images and pseudo-labels. We further investigate the combination method of replay and new data to maximize the feature learning efficiency of the model.

\subsection{Generative Adversarial Networks} 
Current generative models include generative adversarial network (GAN)~\cite{goodfellow2020generative}, variational autoencoder (VAE)~\cite{kingma2013auto}, flow-based generative models~\cite{rezende2015variational}, and diffusion models~\cite{ho2020denoising}. The adversarial relationship between the generator and discriminator in GANs contributes to the creation of visually captivating synthetic images that often possess an astonishing resemblance to real images~\cite{shen2021training}. 
In medical scenes, images shall require high-quality preservation of details and textures. Furthermore, complex structures and textures are easily lost during the multiple-image generation. Considering above challenges, we employ SinGAN~\cite{shaham2019singan} as our pseudo-image generator. SinGAN is designed to learn and capture the underlying distribution of image patches, enabling the generation of diverse and high-quality samples that faithfully preserve the visual content. The architecture of SinGAN consists of a pyramid of fully convolutional GANs, with each GAN focusing on learning the patch distribution at a specific scale of the image. This hierarchical approach empowers SinGAN to produce samples of varying sizes and aspect ratios, exhibiting remarkable diversity while preserving both the overall structure and intricate textures from the original training images.

\section{Methodology}
We define the continual learning sequence with $\mathcal{T}$ tasks, and $t \in \{1,...,\mathcal{T}\}$ means the current task. $\mathcal{D}_t$ represents the training dataset at task $t$, with $x, y$ representing a image and label pair in $\mathcal{D}_t$. ${C}_{old}$ and ${C}_{t}$ denote the classes appearing in previous task $\{1,...,t-1\}$ and current task $t$. Our mini-batch replay method is independent of other regularization/distillation-based methods. We shall first discuss our entropy-based mini-batch pseudo-rehearsal strategy, followed by introducing our SAN-CE loss.

\subsection{Mini-batch pseudo-replay}

\begin{figure*}
    \centering
    \includegraphics[width=0.9\textwidth]{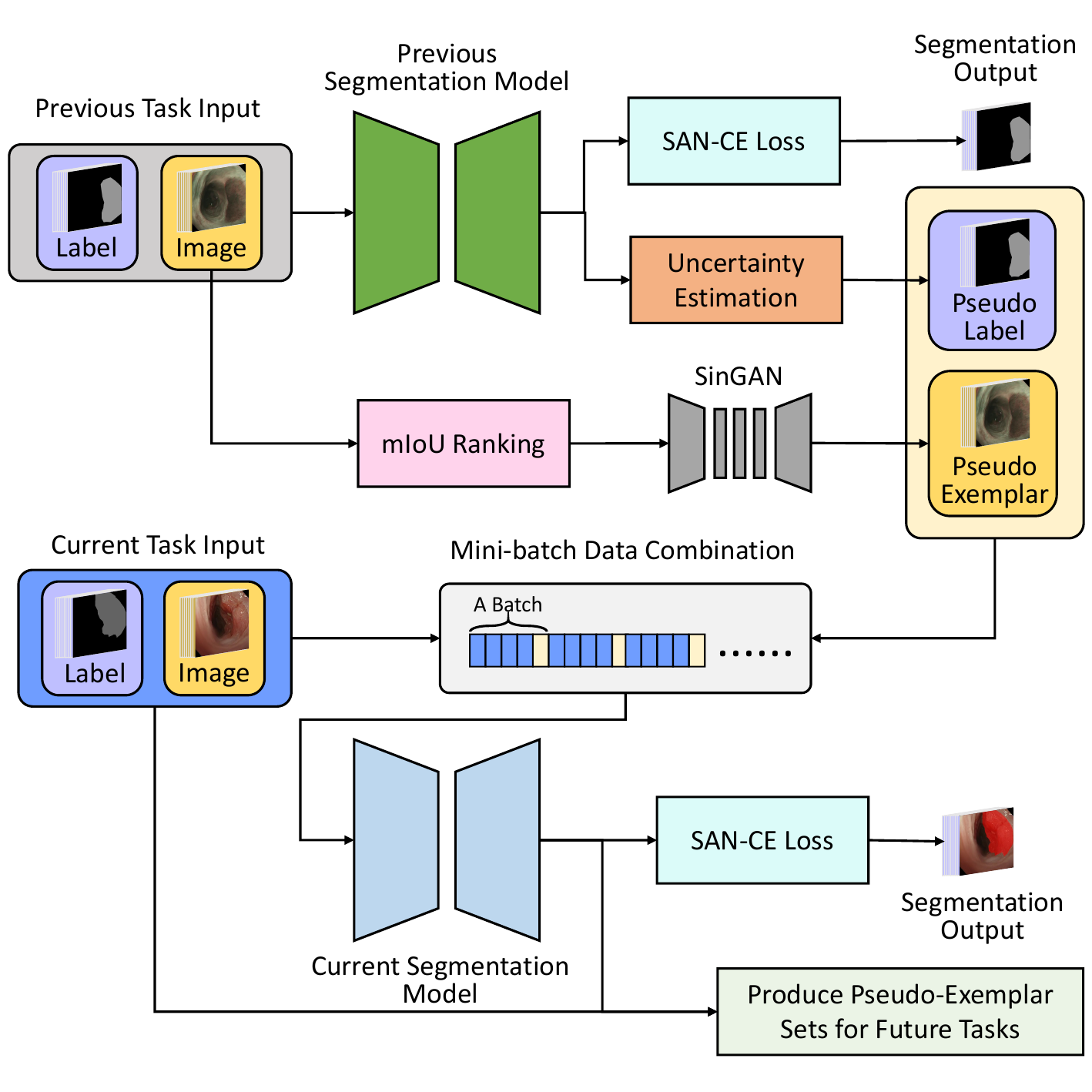}
    \caption{Overview of the proposed EndoCSS framework. We construct the MB-PR strategy by combining the synthetic replay data and current training data at the batch level. SAN-CE loss is employed as the optimization function of our EndoCSS framework. The above procedures will be repeated for Future Tasks $\{t + 1,...,\mathcal{T}\}$.}
    \label{fig:main}
\end{figure*}

We carefully design the procedure of data sample selection and processing. The pseudo-replay and current data shall be combined on the batch level to balance the model bias. The overview can be referred to in Figure~\ref{fig:main}.

\subsubsection{Sample selection with IoU ranking}
As we stated earlier, SinGAN is an unconditional generative model whose input is a single image. The generation process is shown in the lower branch of the Previous Task in Figure~\ref{fig:main}. Considering that train and generate every image is time-consuming, we must strictly filter the training sample sent to SinGAN, and generate reliable pseudo-samples. Besides, samples with high-quality annotations generated by the previous segmentation model are valuable. Works on continual classification have tried to select based on mean feature distance~\cite{rebuffi2017icarl}, plane distance~\cite{chaudhry2018riemannian}, or entropy~\cite{chaudhry2018riemannian}. Nevertheless, we obviously have a better sampling metric in the segmentation tasks, i.e., Intersection over Union (IoU). Firstly, IoU is one of the most significant evaluation metrics for image segmentation. With a high IoU, the image can be considered as the representative of the corresponding class, and the model shall have the excellent predictive ability on it. Secondly, in the current task, the label of the replay data is the pseudo-mask generated from the previous model, and the IoU of the source data in the previous task is the most direct evaluation criterion. If the training result of the source data is excellent, the quality of the generated pseudo-mask shall be improved correspondingly. Therefore, we select the images with the highest IoUs as the training set of SinGAN. In order to address the apparent inter-class imbalance in medical images and ensure unbiased learning, we introduce the ranking of the IoU for each class separately, which allows us to meticulously evaluating and comparing the results distribution. Then, we select a consistent number of images across all classes, mitigating any potential bias and promoting fair.

The above setup ensures we can utilize representative data at the initial stage. We disclose the parameter setup and image number settings in Section~4.

\subsubsection{Entropy-based pseudo-label filtering}
Conditional generative models can serve as an excellent pseudo-replay generator, providing images with class labels in continual classification tasks. However, we need to construct pseudo-replay image-label pairs for CSS tasks. Pseudo-labeling~\cite{lee2013pseudo, wang2022semi, yin2022fishermatch} is a typical strategy in domain transfer for semantic segmentation tasks~\cite{douillard2021plop,vu2019advent}. In this case, we consider generating pseudo segmentation labels using the prediction results from the previous model. 
To measure the pixel uncertainty and discard uncertain pixels in the pseudo-labels, we integrate the entropy-based method to filter pseudo-label with high confidence.
Entropy is a widely used statistical measure of uncertainty. A distribution with low entropy means it is more peaked and less uncertain. We define $p_{i}\in\mathbb{R}^{C}$ as the predicted probabilities from the previous model on the $i$-th pixel. We can achieve the entropy score as follows: 
\begin{equation}
\mathcal{E}\left(p_{i}\right)=-\sum_{c=1}^{C} p_{i}(c) \log p_{i}(c)
\end{equation}

where $C$ refers to the number of classes. Then, the class of each pixel will be set to that with the largest entropy:
\begin{equation}
{y}_{i j}=\left\{\begin{array}{ll}
\arg \max _{c} p_{i}(c) \;, & \text { if } \mathcal{E}\left(p_{i}\right)<\theta \\
\text{eliminate}\;, & \text { otherwise }
\end{array}\right.
\end{equation}
where $\theta$ is the entropy threshold. Therefore, we can easily obtain the pseudo-label with high confidence through entropy, allowing us to establish the generated replay dataset for classes $c \in \mathcal{C}_{1:t-1}$ with accuracy and reliability.

\subsubsection{Mini-batch data combination}
Current replay-based CL methods only naively combine the current dataset and rehearsal exemplars together before feeding the training process~\cite{maracani2021recall,rebuffi2017icarl}. However, naive combination strategies such as concatenating the current and rehearsal dataset may lead to model bias during the learning process~\cite{mai2020batch}.
Therefore, instead of simply mixing the replay data into the current training set, we propose a replay mechanism called mini-batch pseudo-replay (MB-PR).
In the middle of Figure~\ref{fig:main}, the current training batch concatenates a mini-batch of replay samples with a batch of data from the training set. At task $t$, we have a current dataset $\mathcal{D}_t$ and a pseudo-replay set $\mathcal{R}_{1:t-1}$ which contains the pseudo-exemplar set of all previous tasks. Within each batch, we shall define the batch size of current data as $S_\mathcal{D}$ and the batch size of pseudo-exemplars as $S_\mathcal{R}$. We will ensure that each batch of data contains replay data at a fixed ratio based on the amount of data in the entire dataset and exemplar set. Then, each batch extracts certain current data $B_\mathcal{D}$ with the size of $S_\mathcal{D}$ and certain pseudo-exemplars $B_\mathcal{R}$ with the size of $S_\mathcal{R}$, which will be fed into the current model for prediction. The optimizer is then executed using the aggregated batch. The entire training process of the MB-PR mechanism is shown in Algorithm~\ref{alg1}. 

Instead of mixing the replay data directly into the training set, each batch of data is replayed so that both types of data can participate in the model parameter update in gradient descent through the MB-PR process. This strategy enables each batch of data to constrain the model parameters, thus effectively circumventing the risk of losing previous knowledge and experiencing catastrophic forgetting of parameters associated with the previous task. In this case, except for improving the convergence efficiency, the MB-PR strategy also enhances the stability of the training process.

\subsection{Self-adaptive noisy cross-entropy loss}

Literature~\cite{rebuffi2017icarl} has revealed the importance of a stable and robust prediction process in CL. However, the nearest-mean-of-exemplars classification strategy from~\cite{rebuffi2017icarl} is not applicable in CSS because we need to perform pixel-level classification in CSS tasks. Instead, we aim to modulate our loss function to produce a stable and robust training procedure for the predictor.

\begin{algorithm*} 
	\caption{Mini-Batch Pseudo-Replay} 
	\label{alg1} 
	\begin{algorithmic}[1]
        \State \textbf{Train}($\mathcal{D}_t$, $\mathcal{R}_{1:t-1}$, $S_\mathcal{D}$, $S_\mathcal{R}$, $\sigma$) \Comment{$\sigma$ indicates the standard variance of the Gaussian noise}
        \For{epochs}
        \For{batches} 
        \State $\mathcal{D}_t\,\overset{S_\mathcal{D}}{\sim}\,\mathcal{B}_D$  \Comment{Get a data batch of size $S_\mathcal{D}$ from $\mathcal{D}_t$}
        \State $\mathcal{R}_{1:t-1}\,\overset{S_\mathcal{R}}{\sim}\,\mathcal{B}_R$ \Comment{Get a data batch of size $S_\mathcal{R}$ from $\mathcal{R}_{1:t-1}$}
        \State $\mathcal{D}_{train}=\mathrm{Concat}\;[\mathcal{B}_D \; \| \; \mathcal{B}_R]$ \Comment{Concatenate current data batch and replay batch}
        \State $\mathcal{L} = \textbf{SAN-CE Loss}\;(\textit{D}_{train}, \; \sigma, \; S_\mathcal{D} + S_\mathcal{R})$ \Comment{Update the loss}
        \State $\theta \Leftarrow \textbf{Optimizer}\;(\textit{D}_{train}, Loss)$ \Comment{Update the network parameters}
        \EndFor
        \EndFor
        \State \textbf{Return}\, $\theta$
	\end{algorithmic} 
\end{algorithm*}

The cross-entropy loss is the most commonly used loss function in segmentation tasks. For $\mathcal{L}(\textit{x},\textit{y}) = \left\{ \textit{l}_1,...,\textit{l}_\textit{M} \right\}$, the loss of $m$-th sample can be described as: 
\begin{equation}
\textit{l}_\textit{m} = -{\rm log}\frac{{\rm exp}(x_{m,y_m})}{\sum_{c=1}^{C}{\rm exp}(x_{m,c})}\cdot1\left\{y_m \ne {\rm ignore\_index} \right\}
\end{equation}

where $x$ is the input logits of the SoftMax, $y$ is the ground truth, $C$ is the number of classes, $M$ is the number of samples, and ignore\_index is the target value that does not contribute to the input gradient. 

If the output probabilities of $x$ are close to $y$, the sample can be defined as individual saturation. In the model optimization procedure, the optimizer may have fewer chances to move if the sample saturates too early. Thus, the possibility of convergence to a local minima increases~\cite{chen2017noisy} as the number of saturated individuals increases, which will eventually lead to overfitting. Therefore, in regular experiments, an appropriate postponement in sample saturation will help the model to converge globally. In our case, the convergence is challenging due to the complexity of the scenario and the limited amount of data. The segmentation model is always in an underfitted state for most of the samples during the training stage. Therefore, we explore saturated individuals from the opposite perspective to help model fitting Concretely, we promote the sample saturation in underfitting training and increase the contribution of these samples in the back-propagation, giving more chances for the optimizer to move toward global convergence. Based on the above analysis, we explore increasing the logit value of the samples. We use Gaussian noise as the carrier for logit value adjustment. Therefore, the logits fed into SoftMax can be rewritten as:

\begin{equation}
x^G=x[1+{\rm cur\_step}\cdot(\mu+\sigma\left|\xi\right|)]
\end{equation}
where $\mu$ and $\sigma$ are the mean and the standard variance of the Gaussian noise. $\xi$ $\sim$ $\mathcal{N}$(0,1). Here, we set the mean value to 0. cur\_step represents the current task $t \in \{1,...,\mathcal{T}\}$. Therefore, the model can self-adjust the logits according to the current step and get better global convergence. Gaussian noise can further improve the robustness of training effectively. We hereby obtain our self-adaptive noisy cross-entropy (SAN-CE) loss: 
\begin{equation}
\begin{aligned}
\textit{l}_\textit{m} = -{\rm log}\frac{{\rm exp}(x_{m,y_m}[1+{\rm cur\_step}\cdot(\mu+\sigma\left|\xi\right|)])}{\sum_{c=1}^{C}{\rm exp}(x_{m,c}[1+{\rm cur\_step}\cdot(\mu+\sigma\left|\xi\right|)])}\\
\cdot1\left\{y_m \ne {\rm ignore\_index} \right\}
\end{aligned}
\end{equation}

\section{Experiments}

\subsection{Datasets and protocols}
We evaluate our framework on the following two setups.

\textbf{GI Diseases:} Endoscopy Disease Detection and Segmentation Challenge (EDD2020)~\cite{ali2021deep, ali2020endoscopy} is a publicly available endoscopy dataset for GI diseases with 386 frames. It contains five different diseases: Barrett's esophagus (BE), cancer, high-grade dysplasia (HGD), polyp, and suspicious. We divide the training and test set in a 4:1 ratio. Then, we set up and evaluate our method on several CSS protocols, e.g., 4-1, 3-2, and 3-1, respectively. 4-1 means the model will learn 4 classes in the initial step, and then 1 class in the following step ($\mathcal{T}$ = 2 steps). 3-2 means initially 3 then 2 classes (2 steps), and 3-1 means initially 3 classes followed by two times 1 class (3 steps). The class distribution of the EDD2020 dataset can be found in Table~\ref{edd:distribution}. 
 
\begin{table}[!h]
\caption{Class distribution of the EDD2020 dataset~\cite{ali2021deep, ali2020endoscopy}. BE denotes Barrett's oesophagus, and HGD denotes high-grade dysplasia.}
\label{edd:distribution}
\resizebox{0.48\textwidth}{!}{
\centering
\setlength{\tabcolsep}{1mm}{
\begin{tabular}{c|cccccc}
		\hline
		Class & \textcolor{white}{aa} BE \textcolor{white}{aa} & Cancer & \textcolor{white}{aa} HGD \textcolor{white}{aa} & \textcolor{white}{a} Polyp \textcolor{white}{a} & Suspicious\\
		\hline
		Number & 86 & 53 & 74 & 127 & 88\\
		\hline
\end{tabular}}}
\end{table}

\begin{table}[]
\caption{Class distribution of the EndoVis18~\cite{allan2020endovis18} and EndoVis17~\cite{allan2019endovis17} datasets.}
\label{endovis:distribution}
\centering      
\setlength{\tabcolsep}{2.1mm}{
\begin{tabular}{l|c|c}  
            \hline Class Name & EndoVis17 & EndoVis18\\
            \hline Bipolar Forceps & 1127 & 1757 \\
            \hline Prograsp Forceps & 1141 & 972 \\
            \hline Large Needle Driver & 1197 & 278 \\
            \hline Vessel Sealer & 500 & / \\
            \hline Grasping Retractor & 255 & / \\
            \hline Monopolar Curved Scissors & 763 & 1554 \\
            \hline Ultrasound Probe & 454 & 156 \\
            \hline Suction Instrument & / & 272 \\
            \hline Clip Applier & / & 44 \\
		\hline
\end{tabular}}
\end{table}

\begin{table*}[]
\caption{Continual segmentation results on the EDD2020 dataset with different CL protocols. We combine and compare our EndoCSS framework with the current SOTA CSS methods one by one.}
\label{tab:edd}
\centering
\setlength{\tabcolsep}{3.1mm}{
\begin{tabular}{c|ccc|ccc|cccc}
\toprule[1pt]
\multirow{2}{*}{Method} & \multicolumn{3}{c|}{\textbf{4-1} (2 tasks)} & \multicolumn{3}{c|}{\textbf{3-2} (2 tasks)} & \multicolumn{4}{c}{\textbf{3-1} (3 tasks)}                \\ \cline{2-11} 
                        & 0-4 & 5 & All & 0-3 & 4-5 & All & 0-3 & 4 & 5 & All \\ \hline
Fine-tuning             & 18.52    & 32.79  & 20.90   & 26.22    & 13.47    & 21.97   & 18.70    &  0.12  & 30.01  & 17.49   \\\hline \hline
ILT~\cite{michieli2019incremental}  & 16.17    & 19.97  & 16.81   & \textbf{28.07}    & 10.25    & 22.13   & 18.95    &  0.00  & 23.81  & 16.60   \\
ILT w/i EndoCSS          & \textbf{29.95}    & \textbf{25.28}  & \textbf{29.17}   & 26.17    & \textbf{28.08}    & \textbf{26.49}   & \textbf{25.27}    &  0.00  & \textbf{26.36}  & \textbf{21.24}   \\\hline
LWF~\cite{li2017learning}                     & 15.16    & 22.60  & 16.40   & \textbf{25.50}    & 10.19    & 20.39   & 18.10    &  0.00  & 20.89  & 15.55   \\
LWF w/i EndoCSS          & \textbf{20.98}    & \textbf{27.21}  & \textbf{22.02}   & 23.17    & \textbf{30.59}    & \textbf{24.41}   & \textbf{22.98}    &  0.00  & \textbf{21.79}  & \textbf{18.95}   \\\hline
MiB~\cite{cermelli2020modeling}    & 33.76    & 14.27  & 30.51   & \textbf{43.86}    &  0.03    & 29.25   & 11.49    & \textbf{39.39}  & 16.13  & 16.91   \\
MiB w/i EndoCSS          & \textbf{34.12}    & \textbf{29.44}  & \textbf{33.34}   & 37.94    & \textbf{33.07}    & \textbf{37.13}   & \textbf{20.91}    & 37.84  & \textbf{27.85}  & \textbf{24.89}   \\\hline
PLOP~\cite{douillard2021plop}      & 32.57    & 20.98  & 30.64   & 41.23    &  4.18    & 28.88   & 27.25    & 20.71  & 18.48  & 24.70   \\
PLOP w/i EndoCSS         & \textbf{33.94}    & \textbf{33.01}  & \textbf{34.62}   & \textbf{44.59}    & \textbf{12.03}    & \textbf{33.74}   & \textbf{28.33}    & \textbf{53.91}  & \textbf{37.84}  & \textbf{34.18}   \\\hline
RCIL~\cite{zhang2022representation} & 23.63    & 23.22  & 23.56   & 27.34    &  9.12    & 21.27   & 18.09    &  0.00  & 15.90  & 14.71   \\
RCIL w/i EndoCSS         & \textbf{24.66}    & \textbf{24.77}  & \textbf{24.68}   & \textbf{30.96}    & \textbf{26.16}    & \textbf{30.16}   & \textbf{19.89}    & 0.00  & \textbf{25.03}  & \textbf{17.43}   \\\hline \hline
ACS~\cite{memmel2021adversarial} & 24.87    & 27.57  & 25.32   & 31.01    &  \textbf{14.04}    & 25.35   & 18.53    &  44.53  & 27.95  & 24.43   \\
ACS w/i EndoCSS         & \textbf{25.96}    & \textbf{29.01}  & \textbf{26.46}   & \textbf{31.84}    & 13.92    & \textbf{25.87}   & \textbf{20.03}    & \textbf{45.22}  & \textbf{27.98}  & \textbf{25.55}   \\\hline \hline
Joint Training          & 46.54    & 26.83  & 43.26   & 45.30    & 39.18    & 43.26   & 44.85    & 53.32  & 26.83  & 43.26   \\ 
\bottomrule[1pt]
\end{tabular}}
\end{table*}

\begin{figure*}
    \centering
    \includegraphics[width=\textwidth]{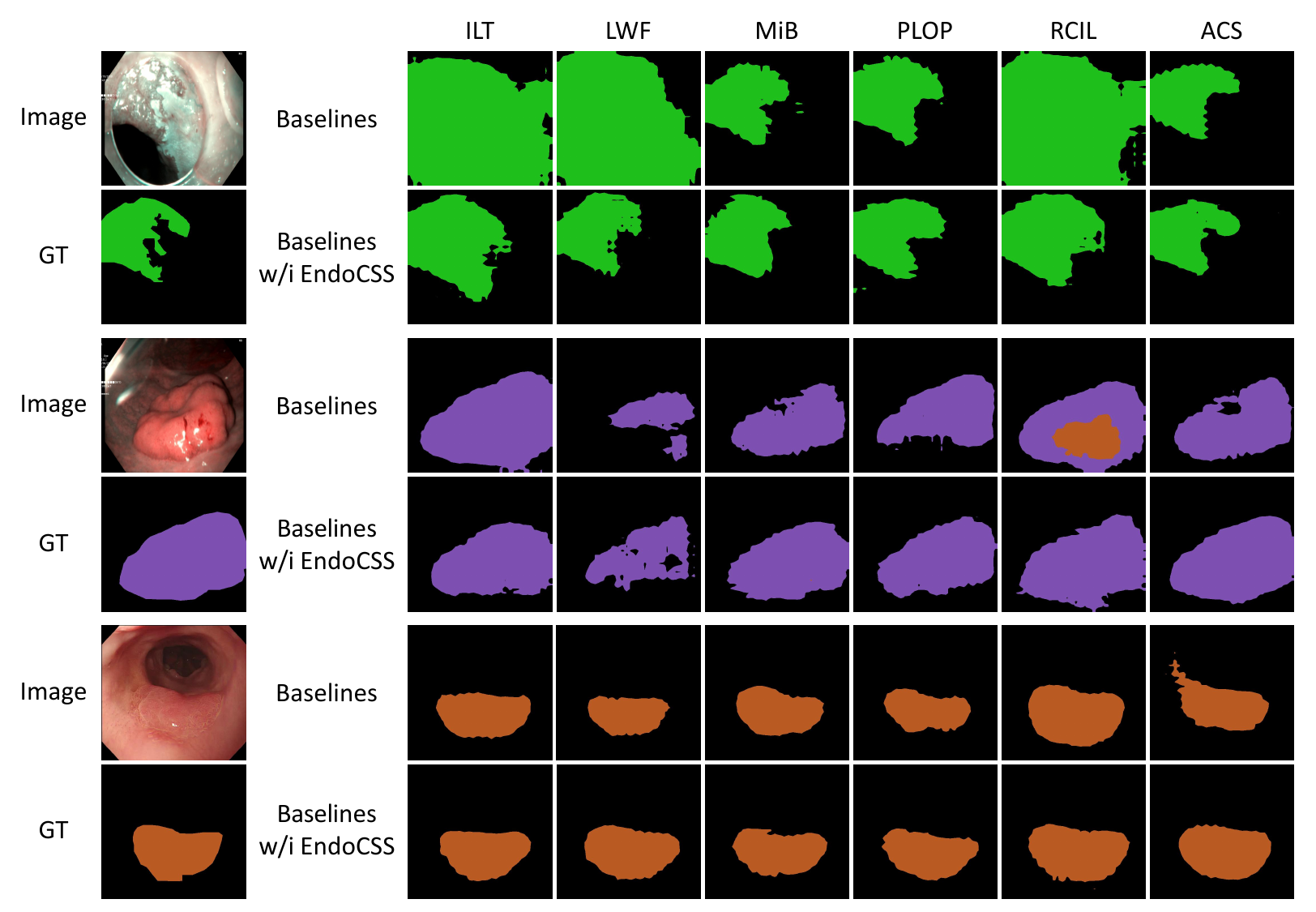}
    \caption{The segmentation visualization of the EDD2020 dataset. The green, purple, and orange indicate cancer, polyp, and high-grade dysplastic.}
    \label{fig:edd_vis}
\end{figure*}

\begin{table*}[]
\caption{Continual segmentation results on the EndoVis18 and EndoVis17 datasets. We set EndoVis2017 as step 0 and EndoVis18 as step 1, with a total of 2 tasks. RC, OC, and NC mean regular/overlapping classes, old non-overlapping classes, and new non-overlapping classes, respectively.}
\label{tab:endovis}
\centering
\setlength{\tabcolsep}{5.7mm}{
\begin{tabular}{c|cccc|c|c}
\toprule[1pt]
& \multicolumn{4}{c|}{EndoVis2017 + EndoVis2018} & EndoVis2017 & EndoVis2018 \\ \cline{2-7} 
\multirow{-2}{*}{Method} & RC     & OC     & NC     & All    & All Classes & All Classes                         \\ \hline
Fine-tuning              & 50.50 & 16.82 &  5.83 & 34.83 & 43.00 & 34.15 \\\hline \hline
ILT~\cite{michieli2019incremental}             & 58.26 & 16.69 &  0.00 & 38.29 & 52.61 & 33.13 \\
ILT  w/i EndoCSS & \textbf{58.83} & \textbf{19.31} &  \textbf{3.63} & \textbf{39.89} & \textbf{52.83} & \textbf{36.66} \\\hline
LWF~\cite{li2017learning}                      & 56.26 & 1566 &  0.00 & 36.89 & 50.43 & 32.11 \\
LWF w/i EndoCSS           & \textbf{57.84} & \textbf{17.33} &  \textbf{0.02} & \textbf{38.17} & \textbf{51.80} & \textbf{33.38} \\\hline
MiB~\cite{cermelli2020modeling}                & 51.38 & \textbf{11.85} &  0.00 & 33.20 & 49.99 & 27.56 \\
MiB w/i EndoCSS     & \textbf{54.05} & 11.05 &  0.00 & \textbf{34.64} & \textbf{51.93} & \textbf{28.93} \\\hline
PLOP~\cite{douillard2021plop}                  & 56.28 & 16.54 &  0.06 & 37.09 & 50.30 & 31.70 \\
PLOP w/i EndoCSS       & \textbf{56.38} & \textbf{17.38} &  \textbf{0.18} & \textbf{37.34} & \textbf{53.73} & \textbf{33.10} \\\hline
RCIL~\cite{zhang2022representation}            & \textbf{56.67} & 13.10 &  0.00 & 36.62 & 52.41 & 32.98 \\
RCIL w/i EndoCSS & 56.53 & \textbf{16.98} &  0.00 & \textbf{37.31} & \textbf{52.77} & \textbf{33.74} \\ \hline \hline
ACS~\cite{memmel2021adversarial}   & 53.32 & 5.94 & 4.36 & 34.05 & 47.90 & 32.72 \\
ACS w/i EndoCSS & \textbf{54.59} &  \textbf{9.15} & \textbf{5.29} & \textbf{35.64} & \textbf{50.61} & \textbf{33.88}\\ \hline \hline
Joint Training & 63.52&	20.56&	00.00&	42.22&	54.30	&40.33\\
\bottomrule[1pt]
\end{tabular}}
\end{table*}

\textbf{Surgical Instruments:} Endoscopy techniques have been widely used on RMIS to provide direct visual signals to the surgeons~\cite{allan2020endovis18,allan2019endovis17}. To further observe the effectiveness of our framework on CSS scenarios with overlapping classes, we further construct a CSS protocol with two publicly accessible datasets on surgical instrument segmentation, i.e., EndoVis18~\cite{allan2020endovis18} and EndoVis17~\cite{allan2019endovis17} dataset. In EndoVis17, we follow~\cite{allan2019endovis17} to divide the 3,000 images into 1,800 as training data and 1,200 as test data. In EndoVis18, we follow~\cite{gonzalez2020isinet} to divide the 2,235 images into 1,639 as training data and 596 as test data. We set EndoVis17 as step $0$, and EndoVis18 as step $1$, with $\mathcal{T}$ = 2 steps. The class distribution of EndoVis17 and EndoVis18 can be found in Table~\ref{endovis:distribution}. We can see there are overlapping and non-overlapping classes in the above two datasets, making the CSS task more challenging.  

\subsection{Implementation details} As we know, rehearsal or pseudo-replay-based CL methods are independent of CL approaches with dynamic architecture or regularization. Therefore, we combine and compare the proposed method with the following state-of-the-art (SOTA) methods one by one: ILT~\cite{michieli2019incremental}, 
LWF~\cite{li2017learning}, MiB~\cite{cermelli2020modeling},  PLOP~\cite{douillard2021plop},  RCIL~\cite{zhang2022representation}, and ACS~\cite{memmel2021adversarial}. Fine-tuning and Joint Training are employed as the upper bound and lower bound of the CSS tasks.

The Deeplab-V3~\cite{chen2017rethinking} with a ResNet101~\cite{he2016deep} backbone pre-trained on ImageNet~\cite{deng2009imagenet} is adopted as our segmentation network. 
All methods are implemented using the Python PyTorch framework on the NVIDIA RTX 3090 GPU.
We use the random horizontal flip and resized crop for data augmentation. 
The batch size is set to 16, and we employ the SGD optimizer.
We adopt the initial learning rate and epoch to $1 \times 10^{-2}$ and $30$ for the first task, and $1 \times 10^{-3}$ and $15$ for the following tasks, respectively.

\subsection{Results}
\subsubsection{Evaluation on EDD2020}
We first evaluate our proposed EndoCSS framework on the EDD2020 dataset. For the replay setting, we set a fixed number of generated pseudo-replay images for each seen class. Here, we define the number of replay images per class as 10. Table~\ref{tab:edd} shows the quantitative experimental result on three different CSS protocols, and Figure~\ref{fig:edd_vis} visualizes the segmentation results on the 4-1 protocol. Overall, our EndoCSS framework presents an overwhelming advantage over all baselines. Specifically, PLOP performs best among all baselines. When we integrate our EndoCSS framework with the PLOP methods, the results further surpass PLOP by 3.98\%, 4.86\%, and 9.48\% mIoU under three different CSS protocols (4-1, 3-2, 3-1). Meanwhile, although the existing SOTA approaches exhibit good retention of old knowledge, they present poor performance on new tasks. Our EndoCSS framework improves this point, not only performing well in mitigating catastrophic forgetting, but also improving the ability of the model to learn new knowledge.
As we state in Section~4.2, Joint Training is regarded as the upper bound of CSS task performance. After our EndoCSS framework is combined with PLOP, it reaches 80.02\%, 77.99\%, and 79.01\% of Joint Training's performance on three CSS protocols, respectively. The results further demonstrate the excellent performance of our proposed framework.

\begin{figure*}
    \centering
    \includegraphics[width=\textwidth]{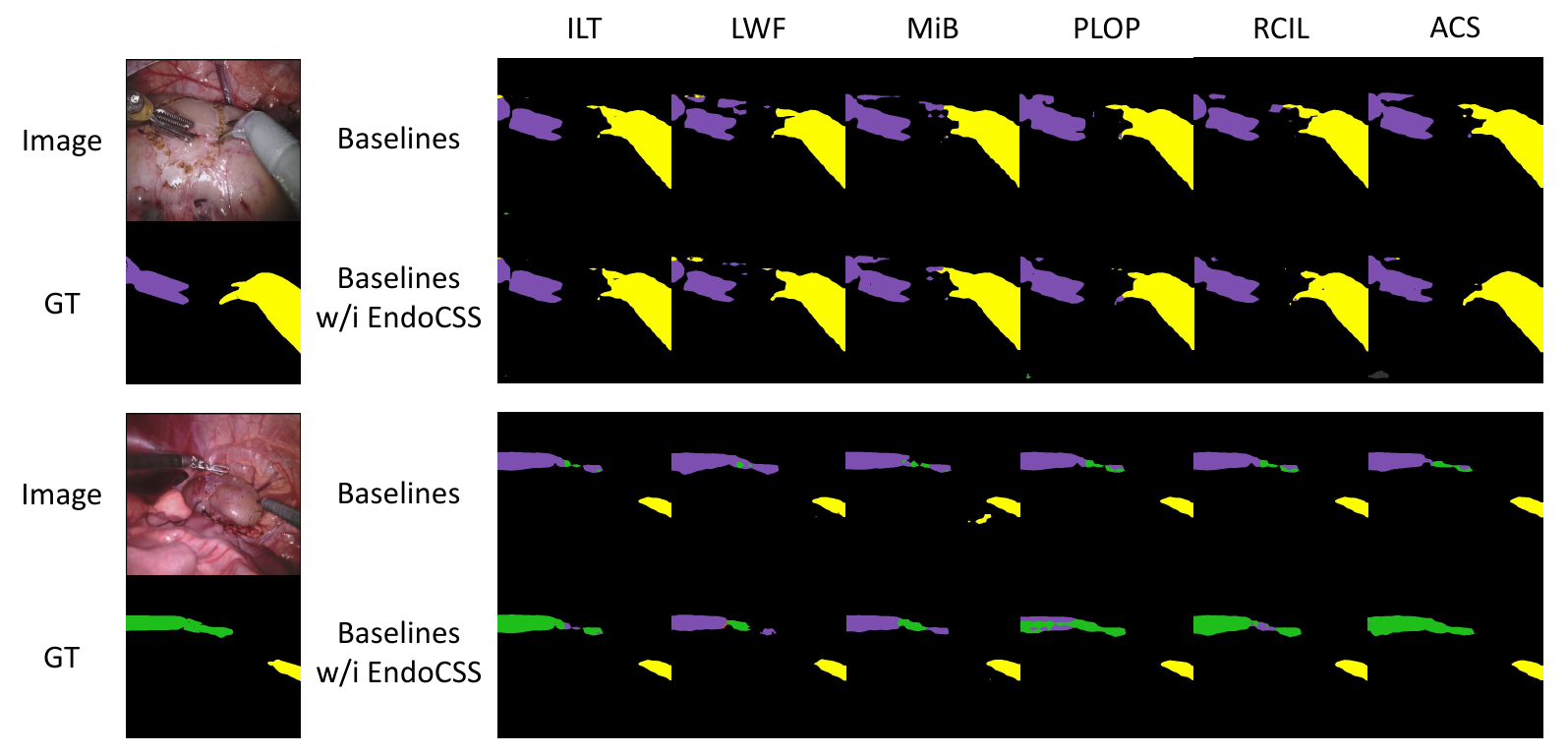}
    \caption{The segmentation visualization of EndoVis datasets. The green, purple, and yellow indicate bipolar forceps, prograsp forceps, and grasping retractor.}
    \label{fig:endovis_vis}
\end{figure*}

\subsubsection{Evaluation on EndoVis18 and EndoVis 17}
We set EndoVis17 as step 0 and EndoVis18 as step 1, with a total of 2 tasks. The CSS on EndoVis17 and EndoVis18 datasets is more challenging and complicated because of the existence of overlapping classes. We define the number of images per class in the previous task as 15, and Table~\ref{tab:endovis} shows the quantitative comparison results. ILT is the best-performing baseline, and our method helps ILT further improve the overall mIoU by 1.60\%. In particular, our framework improves ILT's ability to retain old knowledge and learn new knowledge simultaneously. Moreover, due to the tiny amount of new non-overlapping data in the EndoVis datasets, the original ILT cannot recognize the existence of new non-overlapping classes at all. Our method satisfactorily improves on this and enables ILT to output predictions on new non-overlapping classes. The superior performance of the EndoCSS framework in this challenging scenario further reflects the excellent universality and power of our method. The segmentation results of the EndoVis datasets are visualized in Figure~\ref{fig:endovis_vis}.

\subsubsection{Generative model comparison}

To select a suitable generative model for the pseudo-replay, we compare the generation quality of several similar generative models on the EDD2020 and Endovis2017 datasets. We use Single Image Fréchet Inception Distance (SIFID)~\cite{shaham2019singan} to measure the similarity of the generated images to the original images in terms of features, and use the non-reference metrics Learned Perceptual Image Patch Similarity (LPIPS)~\cite{zhang2018unreasonable} to evaluate the quality of the generated images. The quantitative results in Table~\ref{tab:gan} show that the generated images from SinGAN~\cite{shaham2019singan} possess the best image quality. Therefore, we select SinGAN~\cite{shaham2019singan} as our pseudo-sample generator

\begin{table}[]
\caption{Quality comparison of images produced by different generative models on the EDD2020 and EndoVis17 datasets.}
\label{tab:gan}
\resizebox{0.48\textwidth}{!}{
\setlength{\tabcolsep}{2.2mm}
\begin{tabular}{cccc}
\toprule[1pt]
\multicolumn{1}{l}{}         & Method       & SIDIF $\downarrow$  & LPIPS $\downarrow$  \\ \hline
\multirow{3}{*}{EDD2020}     & SinGAN~\cite{shaham2019singan}       & 160.27 & 0.21   \\
                             & GPNN~\cite{granot2022drop}         & 167.42 & 0.30    \\
                             & SinDiffusion~\cite{wang2022sindiffusion} & 271.06 & 0.46 \\ \hline
\multirow{3}{*}{EndoVis17} & SinGAN~\cite{shaham2019singan}       & 139.89 & 0.22   \\
                             & GPNN~\cite{granot2022drop}         & 170.86 & 0.32   \\
                             & SinDiffusion~\cite{wang2022sindiffusion} & 259.28 & 0.44   \\ 
\bottomrule[1pt]
\end{tabular}}
\end{table}

\begin{table*}[b]
\caption{Ablation study of our EndoCSS framework on the EDD2020 dataset. PR, MB, and SAN denote pseudo-replay, mini-batch combination, and SAN-CE loss. PR represents that we naively mix the pseudo-replay and current data together without the min-batch data combination strategy. We remove the above components one by one and observe the segmentation performance.}
\label{tab:ablation}
\centering
\setlength{\tabcolsep}{3.1mm}
\begin{tabular}{ccc|ccc|ccc|cccc}
\toprule[1pt]
\multirow{2}{*}{PR} & \multirow{2}{*}{MB} & \multirow{2}{*}{SAN} &  \multicolumn{3}{c|}{\textbf{4-1} (2 tasks)} & \multicolumn{3}{c|}{\textbf{3-2} (2 tasks)} & \multicolumn{4}{c}{\textbf{3-1} (3 tasks)} \\ \cline{4-13} 
        &      &  & 0-4 & 5 & All & 0-3 & 4-5 & All & 0-3 & 4 & 5 & All \\ \hline
/& /& /                                       & 32.57 & 20.98 & 30.64 & 41.23 &  4.18 & 28.88 & 27.25 & 20.71 & 18.48 & 24.70 \\
\textbf{\checkmark} & / & /                   & 33.12 & 21.67 & 31.21 & 43.06 &  4.26 & 30.12 & 29.96 & 32.35 &  5.57 & 26.29 \\
\textbf{\checkmark} & \textbf{\checkmark} & / & 33.69 & 22.65 & 31.85 & 43.52 &  5.31 & 30.78 & 30.84 & 30.29 & 13.95 & 27.93 \\
/       & /    & \textbf{\checkmark}          & 32.19 & 30.35 & 31.89 & 41.90 & 12.40 & 32.07 & 24.23 & 53.69 & 32.63 & 30.54 \\
\textbf{\checkmark} & / & \textbf{\checkmark} & 33.53 & 32.04 & 33.28 & 43.91 & \textbf{12.66} & 33.49 & 27.65 & \textbf{55.01} & 34.88 & 33.42 \\                  
\textbf{\checkmark} & \textbf{\checkmark} & \textbf{\checkmark} & \textbf{33.94} & \textbf{33.01}& \textbf{34.62} & \textbf{44.59} & 12.03 & \textbf{33.74} & \textbf{28.33} & 53.91 & \textbf{37.84} & \textbf{34.18} \\                   
 \toprule[1pt]
\end{tabular}
\end{table*}

\subsubsection{Ablation Studies}
Table~\ref{tab:ablation} presents the ablation study to investigate the effect of each proposed component. We decompose our EndoCSS framework into pseudo-replay (PR), mini-batch data combination (MB), and SAN-CE loss (SAN), and then reassemble them to observe the performance. 
Quantitative results show that in the absence of SAN-CE loss, our PR and MB strategies can make the DL model better retain the old knowledge, and also improve the performance of new tasks to a certain extent. Meanwhile, a single SAN-CE loss can significantly improve the performance of new tasks, but some old knowledge is forgotten. 

However, compared with our final results, we can observe the performance degradation of individual components. When we integrate the SAN-CE loss into the MB-PR strategy, our overall performance can be further improved. Therefore, the ablation study demonstrates that our framework can only achieve the best trade-off between old and new tasks when we combine all three components.

\subsubsection{Robustness Experiments}
\begin{figure*}
    \centering
    \includegraphics[width=\textwidth]{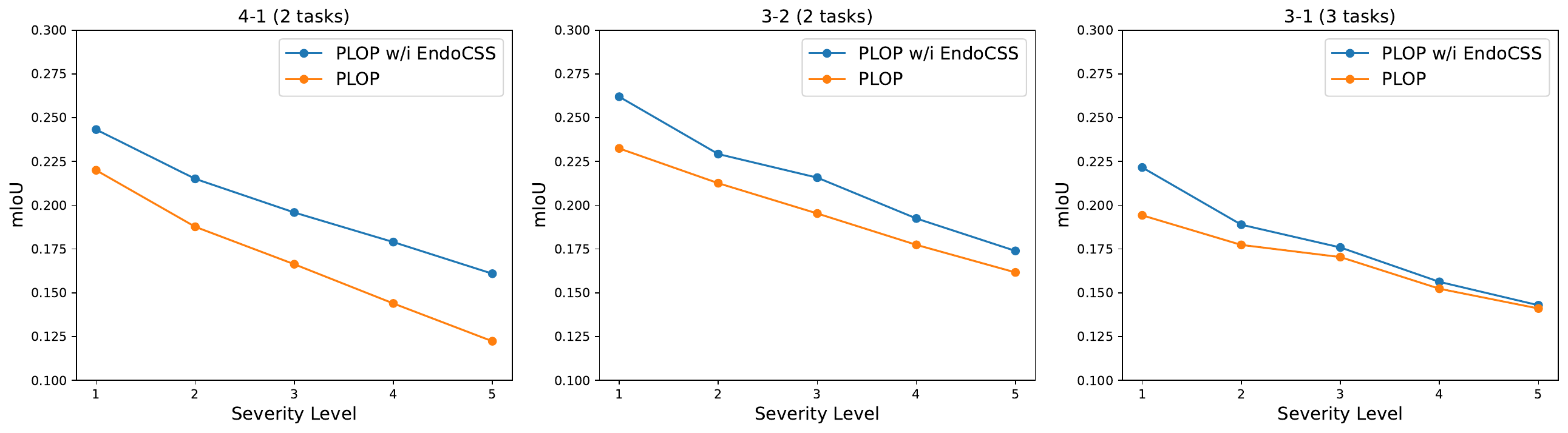}
    \caption{Robustness comparison of our proposed EndoCSS framework against PLOP~\cite{douillard2021plop} on the EDD dataset. We process the data with 18 image corruption approaches at each severity level and average the prediction results.}
    \label{fig:robust}
\end{figure*}

We also conduct a robustness experiment to investigate the performance of our EndoCSS framework against image corruption. Following~\cite{hendrycks2019benchmarking}, we set 5 corruption severity levels and 18 image corruption approaches (defocus blur, glass blur, motion blur, zoom blur, Gaussian blur, contrast adjustment, elastic transform, pixelate, JPEG compression, smoke, brightness adjustment, spatter, saturate adjustment, Gamma correction, Gaussian noise, shot noise, impulse noise, speckle noise) to our test set. We directly perform inference on the corrupted test dataset and average the results of different corruption approaches by severity level. Robustness experiments presented by severity level can be seen in Figure~\ref{fig:robust}. As the corruption severity level increases, the performance of all methods decreases. However, our EndoCSS framework still has a higher mIoU than the baseline, demonstrating the excellent robustness of our framework against image corruption.

\subsubsection{Discussion and limitations}
Overall, our approach has demonstrated significant performance improvements in most comparative experiments. Moreover, as a plug-and-play component, our method is compatible with all existing CSS methods. In visual comparisons, our method exhibits excellent calibration compared to the original approach. It effectively limits predictions to reasonable ranges when encountering unreasonable regions or edges in the original predictions. This is achieved by introducing generated images in the training set, which supplement important representations for the old classes, serving as reminders for the model when learning new tasks. However, we have also observed instances where EndoCSS underperforms compared to the original methods (e.g., the performance of class 0-3 in Task 3-2 of the EDD2020 dataset). Although the introduction of pseudo-replay data allows the current model to simultaneously learn new and old tasks, incorporating more training data increases the number of model iterations and gradually causes the loss of early model weights, leading to the forgetting of earlier tasks. Therefore, a potential future work would be dynamically adjusting the quantity of pseudo-replay images to achieve outstanding adaptive CSS.

The main limitation of our model lies in the high training cost. The model inference procedure of our EndoCSS is similar to the original model, so the inference speed is not affected. However, during training, we need to generate a certain number of replay images using the generative model. This approach effectively addresses privacy concerns and the unavailability of old data in medical imaging. However, the substantial training overhead introduced by the generative model is inevitable. In addition to developing more efficient generative models for training, future work could also explore replay directly from the feature level to address the challenges of training and storage requirements.

\section{Conclusion}

This paper presents a novel continual endoscopic view segmentation learning scheme that learns multiple classes from streaming endoscopic data. The proposed EndoCSS framework can effectively mitigate the catastrophic forgetting problem in endoscopy image segmentation while improving the ability to learn new knowledge. The proposed framework comprises a mini-batch pseudo-replay (MB-PR) strategy and a self-adaptive noisy cross-entropy (SAN-CE) loss. The MB-PR strategy can allow the model to treat replay and current data unbiasedly on the basis of privacy preservation, and perform effective representation learning.
Moreover, the SAN-CE loss can endow the framework with a stable and robust training process.
We conduct extensive comparison, ablation, and robustness experiments on the publicly available EDD2020 dataset. Experimental results demonstrate the superior performance of our method on class-increased endoscopy image segmentation. In addition, the EndoCSS framework also outperforms all baselines on the EndoVis dataset, showing that our framework can handle complex scenarios with overlapping classes. Therefore, our proposed framework is expected to be deployed in clinical real-world streaming scenarios across time or institutions. The DL model can be kept updated by newly appeared data without the issues of limited data storage or data licensing and privacy. In the future, we will further explore the applicability of our method across different medical modalities, scenarios, and tasks.

\printcredits

\section*{Declaration of interest statement}
All the authors declare that there are no interest conflicts that could influence the work presented in this article.
\bibliographystyle{cas-model2-names}

\bibliography{cas-refs}


\end{document}